\documentclass{article}

  \usepackage[preprint,nonatbib]{neurips_2025}

\usepackage[utf8]{inputenc} % allow utf-8 input
\usepackage[T1]{fontenc}    % use 8-bit T1 fonts
\usepackage{url}            % simple URL typesetting
\usepackage{booktabs}       % professional-quality tables
\usepackage{amsfonts}       % blackboard math symbols
\usepackage{nicefrac}       % compact symbols for 1/2, etc.
\usepackage{microtype}      % microtypography
\usepackage{xcolor}         % colors

\usepackage{import}
\usepackage{enumerate}
\usepackage{mathtools}
\usepackage{amsmath}
\usepackage{amsfonts} 
\usepackage{xcolor}

\usepackage{bm}
\usepackage{bbm}
\DeclarePairedDelimiterX{\Esub}[1]{\mathrm{E}}{\big]}{%
	\;\big[\;#1%
}
\DeclarePairedDelimiterX{\Ee}[2]{\mathrm{E}}{\big]}{%
	_{#1}\;\big[\;#2%
}

\DeclarePairedDelimiterX{\infdivx}[2]{(}{)}{%
	#1\;\delimsize\|\;#2%
}
\newcommand{\Xh}{\hat{X}}

\DeclarePairedDelimiterX{\tat}[1]{\theta}{{}}{%
	^{#1}%
}

\newcommand{\defeq}{\vcentcolon=}

\newcommand{\indep}{\perp \!\!\!\!\! \perp}

\newcommand{\real}{\mathbb{R}}

\newcommand{\x}{\mathbf{x}}

\DeclarePairedDelimiterX{\inp}[2]{\langle}{\rangle}{#1, #2}

\newcommand{\bb}{\boldsymbol{\beta}}

\newcommand{\xt}{\tilde{\x}}
\newcommand{\rb}{\bold{r}}

\NewDocumentCommand{\evalat}{sO{\big}mm}{%
	\IfBooleanTF{#1}
	{\mleft. #3 \mright|_{#4}}
	{#3#2|_{#4}}%
}
\NewDocumentCommand{\grad}{e{_^}}{%
	\mathop{}\!% \mathop for good spacing before \nabla
	\nabla
	\IfValueT{#1}{_{\!#1}}% tuck in the subscript
	\IfValueT{#2}{^{#2}}% possible superscript
}
\DeclarePairedDelimiterX{\betat}[1]{{\tilde{\bb}}}{{}}{%
	^{(#1)}%
}

\DeclarePairedDelimiterX{\thetat}[1]{{\tilde{\vtheta}}}{{}}{%
	^{(#1)}%
}

\newcommand{\Xt}{\Tilde{X}}
\newcommand{\Tt}{\Tilde{T}}
\newcommand{\swap}{{\text{swap}(S)}}
\newcommand{\eqd}{\overset{d}{=}}%{=^d}
\def\eqref#1{equation~\ref{#1}}
\def\1{\bm{1}}

\def\rvr{{\mathbf{r}}}
\def\vtheta{{\bm{\theta}}}

\def\mR{{\bm{R}}}

\def\mX{{\bm{X}}}

\DeclareMathAlphabet{\mathsfit}{\encodingdefault}{\sfdefault}{m}{sl}
\SetMathAlphabet{\mathsfit}{bold}{\encodingdefault}{\sfdefault}{bx}{n}
\newcommand{\E}{\mathbb{E}}

\DeclareMathOperator*{\argmin}{arg\,min}

\renewcommand{\P}{\mathbb{P}}

\renewcommand{\E}{\mathbb{E}}

\renewcommand{\vtheta}{\bm{\theta}}
\renewcommand{\rvr}{\mathbf{R}}

\newcommand{\Xb}{\mX}

\usepackage{graphicx}
\usepackage{multirow}
\usepackage{booktabs}
\usepackage{bigstrut}
\usepackage{lipsum}

\usepackage{wrapfig}

\usepackage{thm-restate}
\usepackage{amsthm}

\newtheorem{lemma}{Lemma}
\newtheorem{proposition}{Proposition}

\usepackage[caption=false,font=normalsize,labelfont=sf,textfont=sf]{subfig}
\usepackage{tikz}
\usepackage{verbatim} % Include the verbatim package
\usepackage{algorithmic}
\usepackage{algorithm}
\usepackage[hidelinks]{hyperref} % Include the hyperref package
\usepackage{wrapfig}
\usepackage{multirow}
\usepackage{colortbl}
\usepackage{bigstrut} %used for tables

\usepackage{enumitem}
\newcommand{\Xbr}{\overline{\Xb}}
\renewcommand{\P}{\mathbb{P}}
\newcommand{\Xbh}{\bm{\Xh}}
\newcommand{\Xtb}{\boldsymbol{\Xt}}
\newcommand{\Yb}{\mathbf{y}}
\newcommand{\pe}{pairwise exchangeable }
\newcommand{\pety}{pairwise exchangeability }
\newcommand\norm[1]{\left\lVert#1\right\rVert}

\let\svthefootnote\thefootnote
\newcommand\blankfootnote[1]{%
	\let\thefootnote\relax\footnotetext{#1}%
	\let\thefootnote\svthefootnote%
}

\title{Extending Model-x Framework to Missing Data}

\author{%
 Deniz Koyuncu$^{1}$ 
\quad Alex Gittens $^{1}$ \quad Bülent Yener $^{1}$\\
$^{1}$Department of Computer Science,
  Rensselaer Polytechnic Institute
}

\begin{document}

\maketitle

\begin{abstract}
 One limitation of most statistical/machine learning-based variable selection approaches is their inability to statistically control the selection of uninformative variables. A recently introduced framework, model-X knockoffs \cite{candesPanningGoldModelX2017}, provides False Discovery Rate (FDR) \cite{benjaminiControlFalseDiscovery2001} control to a wide range of models but cannot be used when datasets contain missing values. In this work, we first formulate the dataset shift resulting from improper handling of the missing entries using two commonly used methods: complete-case analysis and missing data imputation. Next, to prevent this distributional shift and maintain the theoretical guarantees of model-X, we introduce sufficient conditions. Specifically, we show imputing the missing variables using the generative model already assumed to be available in the model-X framework, can prevent the dataset shift for a restricted class of missingness mechanisms. Finally, we test the theoretical findings with simulations and investigate how the amount of correlation and missing data rate affected the performance of model-X knockoffs.
\end{abstract}

 \section{Introduction}
Coping with increasing number of variables, optimizing predictive performance, and selecting among candidate scientific hypothesis are reasons motivating the use of feature selection algorithms. Another reality of today's datasets are missing values. Although, there are existing methods for handling the missing values, if applied directly, they can interfere with the assumptions of feature selection algorithms. 

In this work, we will discuss how model-X knockoffs \cite{candesPanningGoldModelX2017}, a new approach in principled feature selection, can be applied to datasets that contain missing values. By principled feature selection we refer to algorithms that aim to identify the Markov Boundary (MB) of a response variable \cite{tsamardinosPrincipledFeatureSelection2003} and identifying the MB is by definition optimal, in the sense that the MB is the smallest subset of variables that is sufficient to describe the conditional distribution of the response variable \cite{margaritisProvablyCorrectFeature}. In addition, model-X knockoffs complements the principled feature selection definition by, under certain regularity conditions, controlling the expected fraction of selected variables which do not belong to the MB \cite{candesPanningGoldModelX2017}.
% Controlling the FDR refers to 
%limiting the variables that are selected due to random chance and is especially important in applications where a selected variable corresponds to a scientific discovery]. 

% Model-x knockoffs framework is intended for principled feature selection as it provide a false selection relevant to MB discovery without making assumptions of the response conditional distribution.

Model-X knockoffs provides a framework for repurposing existing statistical/machine learning feature scorers for MB discovery. 
% It's flexibility follows from the absence of restrictions on the conditional distribution of the response variable. 
When the assumptions of the model-X framework holds, the expected fraction of selections that are conditionally pairwise independent with the response variable is controlled. Formally given explanatory variables $X=(X_1,X_2,\dots, X_d)$ and a response variable $Y$, let $X_{-j}=\{X_l:l\neq j\}$ denotes the variables except $X_j$, then the conditionally pairwise independent variables set is given by $\mathcal{H}_0=\{j:Y\indep X_j \mid X_{-j}\}$ \cite{candesPanningGoldModelX2017}.  The variables in this set are referred to as the null variables. Given the resulting subset $\hat{S}\subseteq \{1,\dots,d\}$ from the feature selection procedure, the FDR \cite{benjaminiControlFalseDiscovery2001} is controlled at a selected level $q$ i.e.  
\begin{equation}\label{eq:fdr}
	\E\bigm[\dfrac{\bigm|\hat{S}\ \cap \ \mathcal{H}_0\bigm|}{\text{max}(1,\bigm|\hat{S}\bigm|)}\bigm]\leq q
\end{equation}
Moreover, under widely used assumptions in the probabilistic graphical models literature, that is equivalent to controlling the expected fraction of selections that are outside of the MB of the response \cite{candesPanningGoldModelX2017}.

The key advantage of model-X knockoffs is that it achieves the FDR control implied by \eqref{eq:fdr} without making any restrictions on the response conditional distribution $\P_{Y\mid X}$. This advantage relies on an accurate generative model $\P_{X}$ which is used to sample a special synthetic design matrix statistically identical to the original design matrix. This procedure, by contrasting the scores feature selection algorithms attribute to the variables in the original and the synthetic design matrices, enables filtering the potentially null variables.

%and filtering the potentially null variables.

%These design matrices for calibrating the feature 

%Model-X achieve the FDR control implied by equation \eqref{eq:fdr} in three steps. First, through an accurate model a statistically identical “knockoff” is sampled for each variable. Second, using the original training set, and the newly created synthetic dataset consisting of the knockoffs, the importance of each original variable and its knockoff are fairly assessed. Third, the resulting scores are calibrated to filter the null variables. 
%[check the arxiv version intro as well, give the intuition]
%model-x has been applied such as the genome-wise association studies tend to 

%Model-X requires special consideration because of its rather unorthodox methodology which involves generating a synthetic design matrix to calibrate feature selection algorithms.

While model-X is a flexible framework, its original formulation \cite{candesPanningGoldModelX2017} assumes the design matrix has been completely observed. However, datasets in practice can contain missing entries which can interfere with assumptions of model-X. The primary mechanism model-X can get affected by is distribution shifts resulting from the missing data. Firstly, shifts in the response conditional distribution can change the reference null variable set. For example, important variables missing in some examples can lead to spurious correlations and undercount of false discoveries. Secondly, due to shifts in the explanatory variable distribution, an originally accurate generative model might fail to capture the distribution resulting from missing data. 

To our knowledge, such interactions between missing data and model-X have not been considered in the literature. Instead, earlier works applying the model-X reverted to either using potentially biased missing data imputation strategies \cite{candesPanningGoldModelX2017, masudUtilizingMachineLearning2021} or omitting the variables with missing entries from the analysis altogether \cite{jiangKnockoffBoostedTree2021,fuHighdimensionalVariableSelection2021}. As the model-X framework is increasingly used in applications where missing data occurs naturally (e.g. genome-wide association studies), it is important to assess the conditions when missing data handling methods can preserve its statistical guarantees. This work will study the subtle interactions between missing data analysis and model-X knockoffs, specifically in the context of Complete-Case Analysis (CCA) and missing data imputation.

In Section \ref{sec:b_background}, we give the necessary background regarding model-X knockoffs and missing data handling methods. In Section \ref{a:sec:main}, we introduce the sufficient conditions and strategies for preserving the theoretical guarantees of model-X. In Section \ref{a:sec:results}, simulations will be used to demonstrate our theoretical findings. In Section \ref{a:sec:rel} we review the existing literature and conclude. 

Our contributions in this work are to

\begin{itemize}
%	\item Formulating the effects of missing data handling methods, specifically CCA and imputation, on the assumptions of the model-X framework
%	\item Formulate the effects of CCA and missing data imputation on the model-X framework, specifically, its reference null hypothesis set and generative model, and to
%	when the missing entries are handled using CCA and imputation, null hypothesis set model-X references and the underlying generative model can change,	
%	\item Introduce a missing data imputation strategy which repurposes the model-X generative model and preserves the original data distribution for a class of missingness mechanisms.
\item Identify the two main ways missing data handling methods CCA and imputation can invalidate the FDR control guarantees of the model-X framework, namely by removing \pety and changing the null variable set,
\item Introduced the Missingness with Ignorable Response (MIR) condition to 
determine the natural conditions under which imputing using the generative model, $\P_X$, is sufficient to hold all assumptions of model-X knockoffs in the imputed data,

%\item Determine the natural conditions under which imputing using the generative model, $\P_X$, is sufficient to hold all assumptions of model-X knockoffs in the imputed data,
%\item One of the components o
%	\todo[inline]{mention you introduce and justify the MIR condition}
%	\item Proving posterior sampled imputation allows reusing existing knockoffs samplers
%	\item Proposing an univariate method that doesn't require posterior sampling
%	\item Showing knockoff sampling can be jointly done with imputation
	% \item Quantifying the joint effect of correlation and missing data on power
	% \item Describing the trade-off between the number of missing data and observations
\end{itemize}

% The imputation method estimates the missing variables $X_{m}^{(i)}$ using the observed variables $X_{o}^{(i)}$ where $o\subseteq\{1,\dots,p\}$ and $m=\{1,\dots,p\}\setminus o$ represents the indices of the observed and missing R.Vs. Formally, the estimate is denoted as $\hat{X}_{m}^{(i)}= g(X_{o}^{(i)})$ where $g(\cdot):\real^{|o|}\rightarrow \real^{|m|}$ is the prediction function. The distribution of $X_o,\hat{X}_m)^T$ denote the imputed one.

% If only a subset of the variables in the i'th row denoted with $o^{(i)}\subseteq\{1,\dots,p\}$ are observed, 
% \{1,\dots,p\}\setminus o$ of the variables in the i'th row is 

% Imputation problem can be framed as a prediction problem where an estimate of the missing values $\hat{X}_m$ is given by

% $\xt \sim \mathcal{A}(x_c,P_X) 

% Model-x knockoffs framework 

%\subsection{Notation}
% Variable selection algorithms are used for coping with
\begin{table}[h]
	\centering
	\caption{Notation used and their explanations.}
	\resizebox{1\columnwidth}{!}{%
		% Table generated by Excel2LaTeX from sheet 'Chapter A'
\begin{tabular}{r|l}
\hline
\multicolumn{1}{|l|}{Notation} & \multicolumn{1}{l|}{Explanation} \bigstrut\\
\hline
$X$   & Random vector of the modeled variables \bigstrut[t]\\
$\x$  & Realization of $X$ \\
$R$   & Random vector of the missingness indicators \\
$\rb$ & Realization of $R$, a missingness pattern \\
$o$   & Indices of the observed variables in $\rb$ \\
$m$   & Indices of the missing variables in $\rb$ \\
$\P_X$ & Joint distribution of the modeled variables \\
$\P_{R|X}$ & Conditional distribution of $R$ given $X$, the missingness mechanism \\
$Y$   & Response random variable \\
$\hat{S}$ & Subset of features selected by model-X \\
$\mathcal{H}_0$ & Set of null variables \\
$\P_{Y\mid X}$ & Conditional distribution of $Y$ given $X$, the response conditional distribution \\
$\Xt$ & Random vector of the knockoffs \\
$\Xb$ & $N\times d$ matrix each row a realization of $X$ \\
$\Xtb$ & $N\times d$ matrix each row a realization of $\Xt$ \\
$\mR$ & $N\times d$ matrix each row a realization of $R$ \\
$\Xbr$ & $N\times d$ matrix, the partially observed $\Xb$ \\
$\Yb$ & N-dimensional vector each element a realization of $Y$ \\
$\eqd$ & Equality in distribution \\
$\swap$ & Swap operator see Section \ref{sec:b_background}. \\
$P_{\Xt\mid X}$ & Conditional distribution of $\Xt$ given $X$, the knockoff sampler \\
$t(\cdot,\cdot)$ & Scoring function \\
$T$   & d-dimensional vector, scores of the originals \\
$\Tt$ & d-dimensional vector, scores of the knockoffs \\
$\tau$ & Filtering threshold used in model-X \\
$q$   & False discovery rate threshold \\
$Q_{X_m\mid X_o}$ & A priori known imputation model \\
$\mathcal{M}$ & Set of features corresponding to the Markov Blanket of $Y$ \\
$S^*$ & Set of non-null features used in the simulations \\
$\mathcal{R}$ & Set of features permitted to have missing data \\
$\rho$ & Correlation strength \\
$p_0$ & Missingness rate \\
\end{tabular}%

	}
	\label{tab:addlabel}%
\end{table}%

%The random vector of d explanatory variables is denoted as $X=(X_1,X_2,\dots,X_d)$, its realization as $\x\in\ref{xx}$ and i.i.d. realizations are stacked as a matrix $\Xb\in\real^{N\times d}$. The response random variable is denoted as $Y$, its realization as $y\in\real$, and i.i.d. realizations are stacked as a vector $\Yb\in\real^N$. An additional random vector $\Xt=(\Xt_1,\Xt_2,\dots,\Xt_d)$ is used to indicate the ``knockoffs'' and its i.i.d. realizations are denoted as matrix $\Xtb\in\real^{N\times d}$. The random vector determining which variables are missing is denoted as $R=(R_1,R_2,\dots,R_d)$, its realization $\rb\in\{0,1\}^d$ and in the matrix form $\mR\in\real^{N\times d}$. The feature scorer $t([\Xb,\Xtb],\Yb)$ takes in a concatenated matrix of size $N\times 2d$ and the response vector, then outputs a vector of size $2d$. Given a set $\mathcal{M}$, its complement is denoted as $\overline{\mathcal{M}}$.

\section{Background}\label{sec:b_background}

In the knockoffs methodology, given an $N\times d$ input design matrix $\Xb$ an additional $N\times d$ matrix (called knockoffs) $\Xtb$ is sampled such the two matrices are pairwise exchangeable i.e.

\begin{equation}
	(\Xb,\Xtb)_\swap\eqd \Xb,\Xtb
\end{equation}

holds for all $S\subseteq\{1,\dots,d\}$ \cite{candesPanningGoldModelX2017}. The symbol $\eqd$ denotes equality in distribution and the $\swap$ operator, given two matrices, swaps the columns indexed by $S$ in the first matrix with the ones in the second and vice versa. For that condition to hold, assuming rows of the original design matrix $\Xb$ are independent and identically distributed (i.i.d.) \footnote{If the i.i.d. assumption does not hold, knockoff sampling becomes more challenging.} samples from the random vector $X$, each row of $\Xtb$ is 
generated as i.i.d. samples from an additional random vector $\Xt=(\Xt_1,\Xt_2,\dots, \Xt_d)$ to ensure $X$ and $\Xt$ are pairwise exchangeable. This is achieved by sampling $\Xt$ conditioning on $X$ i.e. $\Xt \mid X \sim P_{\Xt\mid X}(\cdot;\Xt)$ and tailoring the knockoff sampler, $P_{\Xt\mid X}$, to the underlying generative model $\P_X$ for satisfying \pety.

%Formally, the i'th row of the knockoff matrix, 
%$\xt^{(i)}$ is sampled conditioning on i'th row of the original design matrix, $\x^{(i)}$, i.e. $\xt^{(i)} \mid \x^{(i)} \sim P_{\Xt\mid X}(\cdot;\x^{(i)})$ using a knockoff sampler and this process requires the underlying generative model $\P_X$ to ensure \pety.

%. This conditional distribution $P_{\Xt\mid X}$ uses the underlying generative model $\P_X$ to ensure originals $X$ and knockoffs $\Xt=(\Xt_1,\dots, \Xt_d)$ are pairwise exchangeable. 

%Intutiavely consider a store analogy

In the next step the concatenated $N\times 2d$ matrix, $[\Xb,\Xtb]$, and the response, $\Yb$, are given to a statistical learning algorithm that assigns weights to each of the $2d$ columns, i.e,
%T,\Tt = 
\begin{equation}
	\underbracket{(T_1,\dots,T_d)}_{T},\underbracket{(\Tt_1,\dots,\Tt_d)}_{\Tt}=t([\Xb,\Xtb],\Yb)
\end{equation}

The only constraint for the statistical learning algorithm is that whenever a pair of columns are swapped, the output weights should also be swapped accordingly \cite{candesPanningGoldModelX2017} i.e.,  
\begin{equation}
	t([\Xb,\Xtb]_\swap,\Yb)=(T,\Tt)_\swap
\end{equation}
holds for $S\subseteq\{1,\dots,d\}$. Next, a knockoffs filter is applied to determine a calibrated threshold $\tau$ as follows \cite{candesPanningGoldModelX2017}:
% (which we refer as knockoffs) 

\begin{equation}
	\normalfont \tau \defeq \text{min}\bigm\{t>0:\dfrac{1+\bigm|\{j:T_j-\Tt_j\leq -t\}\bigm|}{\bigm|\{j:T_j-\Tt_j\geq t\}\bigm|}\leq q\bigm\}
\end{equation}
Finally, the set of features whose original weights are sufficiently higher than their knockoffs are selected i.e. $\hat{S}=\{j:T_j-\Tt_j>\tau\}$. This threshold requires a selected feature’s score to exceed the score of its corresponding knockoff but it also ensures the gap is significant enough by comparing how many times this gap can be achieved accidentally. If the \pety of $\Xb$ and $\Xtb$ is satisfied, the rows of the $[\Xb, \Xtb]$ are independent and the statistical learning algorithm $t(\cdot)$ satisfies the swap condition, then the FDR is controlled at a pre-selected level $q$ as denoted in Eq. \ref{eq:fdr} \cite{candesPanningGoldModelX2017}.

Keeping aside the swap condition for the statistical learning algorithm, the \pe condition and the row independence condition has to be justified for the knockoff sampler used in practice.

% $\mathcal{H}_0$
\subsection{Effects of Missing Data Under Complete-Case Analysis}
\begin{figure}
	\centering
	\includegraphics[width=1\linewidth]{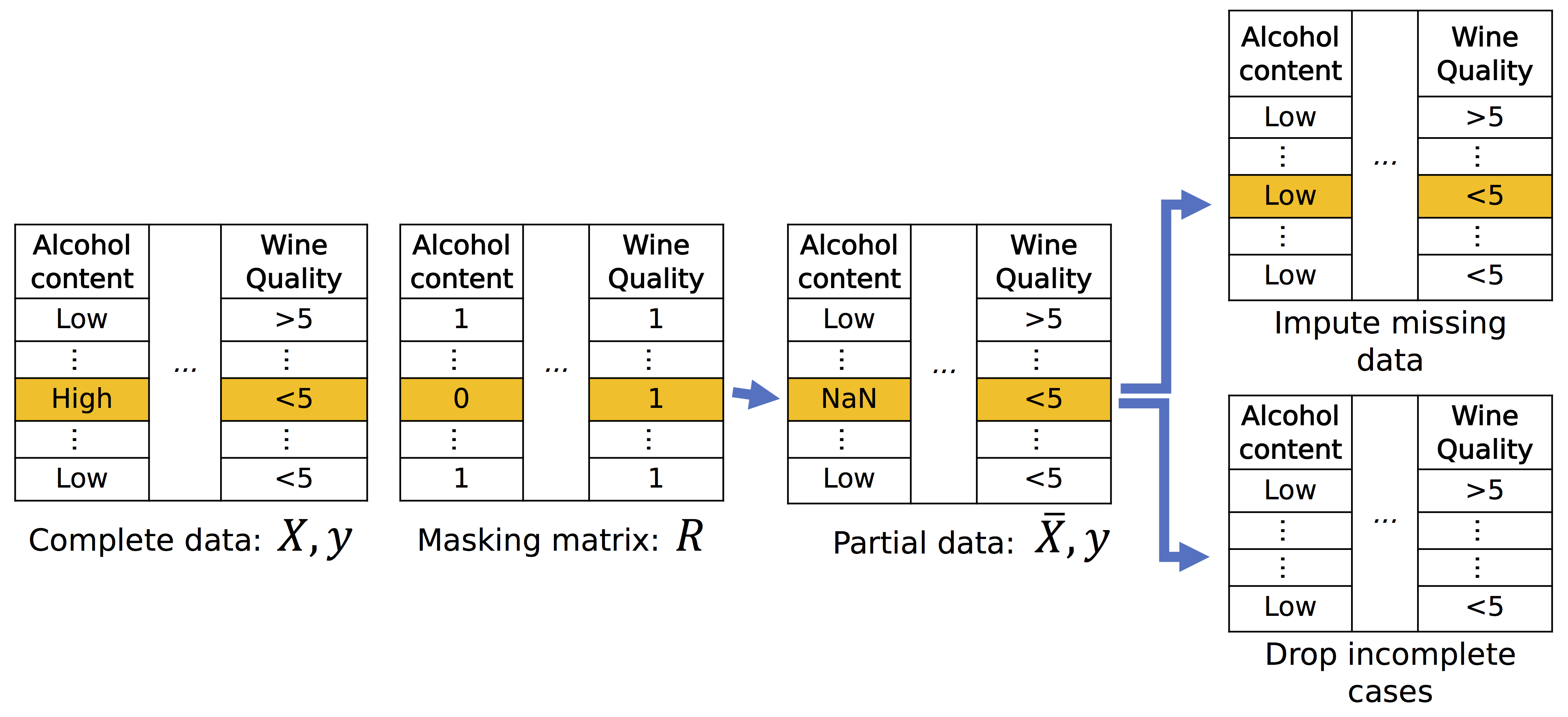}
	\caption{The missing data analysis pipeline under missing data imputation and complete-case analysis.}
	\label{fig:overviewforthesiswithlabels}
\end{figure}

%To the best of our knowledge, the effects of mi
%The model-X framework is introduced thinking the dataset is completely observed while datasets in many practical applications (e.g. xx, yy) can contain missing values. Our goal in this paper is to understand where missing data fits in the model-X framework.
Model-X was introduced considering completely observed datasets but practical datasets may contain missing entries. It is important to understand how the conditions model-X relies on for preserving the FDR can be effected in the missing data setting. Given a completely observed matrix $\Xb$ and a missingness indicator matrix
$\mR\in\{0,1\}^{N\times d}$, we denote a partially observed matrix as $\Xbr$ such that $\Xbr_{i,j}=\text{NA}$ if $\mR_{i,j}=0$ while $\Xbr_{i,j}=\Xb_{i,j}$ if $\mR_{i,j}=1$. Here, we restrict the missingness to the explanatory variables and assume $\Yb$ is completely observed. Our analysis also treats different rows as i.i.d. samples from the distribution $\P_{X,Y,R}$ where $R=(R_1,R_2,\dots,R_d)$ denotes the binary missingness indicators such that $R_j=1$ indicates the corresponding variable $X_j$ is observed and $R_j=0$ indicates $X_j$ is missing. 

%Formally, in this setting each modeled variable $X_j$ has a corresponding The conditional distribution of all missingness indicators, $R=(R_1,R_2,\dots,R_d)$, 

A straightforward approach for using model-X in the partially observed setting is to discard the rows with missing data and apply the analysis (e.g. knockoff sampling) to the completely observed rows of $\Xbr$ and their corresponding responses $\Yb$. This procedure is referred to as CCA in the statistics literature \cite{littleStatisticalAnalysisMissing2002}. Even this simple procedure effectively changes the distribution of the input dataset because in the remaining rows, the joint distribution of the  explanatory variables and the response changes from $\P_{X,Y}$ to the distribution conditioned on the missingness pattern being all observed $\P_{X,Y\mid R}(\cdot,\cdot\mid \mathbf{1})$. 

%the distribution of the completely observed rows is conditioned on the event the missingness pattern random vector equals to one, i.e., $\{R=\mathbf{1}\}$. Subsequently, 

This distribution shift can have two effects. First, because the explanatory variable distribution is shifted to $\P_{X\mid R}(\cdot\mid \mathbf{1})$, a knockoff sampler designed for $\P_{X}$ might not satisfy the pairwise exchangeability. Second, and more importantly, because the response conditional distribution changes to $\P_{Y\mid X, R}(\cdot,\mid \cdot, \mathbf{1})$, the null hypothesis set also changes. Under this conditional distribution, the new null hypothesis set $\mathcal{H}_0^{(cca)}\subseteq\{1,\dots,d\}$ contains the $j$'th explanatory variable if $X_j$ is conditionally independent of the response $Y$ given both the remaining explanatory variables, $X_{-j}$, and the event $\{R=\mathbf{1}\}$, i.e., 
\begin{equation}P_{Y\mid X_j,X_{-j},R}(y\mid \x_j,\x_{-j},\mathbf{1})=P_{Y\mid X_{-j},R}(y\mid \x_{-j},\mathbf{1})
\end{equation}
holds for all $(\x,y)\in \mathcal{X}\times\mathcal{Y}$, such that $\P_{X,R}(\x,\mathbf{1})>0$. 

In general, the new null hypothesis set is not equals to the original null hypothesis set, i.e., $\mathcal{H}_0^{(cca)}\neq \mathcal{H}_0$ and due to these two effects, using knockoffs with CCA does not guarantee FDR control.
%\[\]
%to a definition which includes an additional condition that all explanatory variables are observed, i.e., $\{R=\mathbf{1}\}$,
%\[\mathcal{H}_0^{(cca)}=\{j:Y\indep X_j \mid X_{-j}, \{R=\mathbf{1}\}\}\]

%where $Y\indep X_j \mid X_{-j}, \{\R=\mathbf{1}\}$ holds if
%\[P_{Y\indep X_j, X_{-j}, R}()\]
% \ref{xx}.

%Violating these two conditions, FDR control to the original set of null hypothesis would not be possible in general. 

%Even it MCAR holds still discarding complete rows not efficient]
%xx can change. 

%can shift i.e. $\P_{X\mid R=\mathbf{1}}$ and 
\subsection{Effects of Missing Data Under Imputation}
As an alternative to discarding missing rows, one could impute the missing entries first and apply model-X knockoffs on the resulting imputed matrix. In this setting, the assumptions of the model-X framework have to be verified with respect to the to joint distribution of the imputed matrix $\Xbh \in \real^{N\times d}$ and the response vector $\Yb$. Consider the case that the imputed matrix results from an imputation process, denoted as $g$, which takes in the partially observed data and the response labels i.e.
\begin{equation}
	\Xbh = g(\Xbr,\Yb).
\end{equation}
Commonly used methods such as matrix completion algorithms \cite{mazumderSpectralRegularizationAlgorithms2010}, and the sequential imputation method Multiple Imputation by Chained Equations (MICE), \cite{van1999flexible} are concrete instantiations of the black-box $g(\cdot)$. In this setting, if the imputation model is inexact, it can create a distributional shift and, similar to CCA, invalidate the guarantees of the FDR control. In addition, because an imputation algorithm uses the observed entries of a row to impute the missing entries of a different row (e.g, consider mean imputation), this process can break the independence between separate rows.

%\clearpage
%the set of null variables can 
%indicate the entries of $\Xb$ to be made missing with the binary masking matrix $\mR\in\{0,1\}^{N\times d}$. After selecting the masking matrix, the adversary hides the corresponding entries of the original matrix to obtain the matrix $\Xbr$ of partially observed data: $\Xbr_{i,j}=\text{NA}$ if $\mR_{i,j}=0$ while $\Xbr_{i,j}=\Xb_{i,j}$ if $\mR_{i,j}=1$. 
%in this case the input will be an matrix with imputed missing values $\bold{\Xh}$ and model-X's assumptions has to be with respect to the joint distribution of  $\bold{\Xh}$ and $\Yb$. [breaks the indepence assumption][otherwise using response is not feasible ][]
% but still the imputation process can result in a distributional shift
When the imputation model is known \emph{a priori} or estimated from an independent dataset, each row can be imputed independently. Yet, still, a mismatched imputation model can invalidate the assumptions of model-X knockoffs. Accordingly, let $\Xh=(\Xh_1,\Xh_2,\dots,\Xh_d)$ denote the explanatory variables resulting from the imputation and given a missingness pattern $\rb\in\{0,1\}^d$, $m=\{j:\rb_j=0\}$ and $o=\{j:\rb_j=1\}$ denote the indices of the missing and the observed variables corresponding to the pattern respectively. Consider an imputation process which for each missingness pattern $\rb$, keeps the observed explanatory variables intact $\Xh_o=X_o$ and, assuming $m\neq\emptyset$, imputes the missing ones, $\Xh_m $, with \emph{a priori} fixed imputation model using the observed explanatory variables $X_o$. For a particular missingness pattern $\rb$, this imputation process, without loss of generality, can be expressed as sampling the missing variables, $\Xh_m$, from a conditional distribution $Q_{X_m\mid X_o}$ using the observed ones. In turn, the resulting gap between the joint distribution resulting from imputation, $\P_{\Xh,Y}$ and the original joint distribution, $\P_{X,Y}$, can be expressed by the following weighted sum over the different missingness patterns as follows:
%To illustrate the effect of missing values imputation on those two assumptions we define the missing data problem next. 
%Model-x assumes 
%Let $R=(R_1,\dots,R_p)^T$ denote the missing value indicator R.Vs. $R_j$ takes values in $\{0,1\}$ and $R_j=1$ denotes $X_j$ is missing. Let $m=\{j:R_j=1\}$ and $o=\{1,\dots,p\}\setminus m$ denote the sets of missing and observed values. Given a set $S\subseteq\{1,\dots,p\}$, $X_S=\{X_j:j\in S\}$ denotes the R.Vs indexed by set S. We denote the N i.i.d. observations and masks as $\{X^{(i)}_{o^{i}},R^{(i)}\}_{i=1}^N$ with varying missing values indices $\{m^{i},o^{i}\}_{i=1}^N$.
%To understand the effect of imputation on the knockoffs framework, we classified imputation algorithms into two. In the first category, all observed samples are used to impute all missing values jointly as follows:
%\[(\hat{X}_{m^1}^{(1)},\dots, \hat{X}_{m^N}^{(N)}) = g(X_{o^1}^{(1)},\dots,X_{o^N}^{(N)})\]
%This implicitly assumes the imputation function is assumed to be known apriori (or estimated from a different dataset). But the problem is for an arbitrary $Q_{\Xh_m\mid X_o}$ the resulting distribution of the imputed vector $P_{\Xh^{(i)}}$ is no longer equal to the original $P_{X^{(i)}}$. Dropping the superscripts for brevity, we illustrate this in the Eq. \ref{eq:back}.
%   &=\sum_{r}P_{\Xh,R}(x,r)-P_{X,R}(x,r)\\
\begin{align*}
	&\P_{\hat{X},Y}(\x,y)-\P_{X,Y}(\x,y)=\sum_{\rb\in\{0,1\}^d}\P_R(\rb)\left[\P_{\Xh,Y\mid R}(\x,y\mid \rb)-\P_{X,Y\mid R}(\x,y\mid \rb)\right]\stepcounter{equation}\tag{\theequation}\\
	\intertext{Notice when $\rb=\mathbf{1}$, the data distribution is preserved and the summation is effectively over all $\rb\neq\mathbf{1}$.}
	&=\sum_{\rb\neq\mathbf{1}}\P_R(\rb)\left[\P_{\Xh,Y\mid R}(\x,y\mid \rb)-\P_{X,Y\mid R}(\x,y\mid \rb)\right]\stepcounter{equation}\tag{\theequation}
	\intertext{Applying the chain rule and observing that the observed explanatory variables have the same conditional distribution in the original set and the imputed set, ${\P_{X_o,Y\mid R}=\P_{\hat{X}_o,Y\mid R}}$, the equation can be rearranged as}
	&=\sum_{\rb\neq\mathbf{1}}\P_R(\rb)\P_{X_o,Y\mid R}(\x_o,y\mid  \rb)\left[\P_{\Xh_m\mid \Xh_o,Y,R}(\x_m\mid  \x_o,y,\rb)-\P_{X_m\mid X_o,Y,R}(\x_m\mid  \x_o,y,\rb)\right]\stepcounter{equation}\tag{\theequation}\\
	\intertext{Plugging in the imputation distribution, $Q_{X_m\mid X_o}$, results in}
	&=\sum_{\rb\neq\mathbf{1}}\P_R(\rb)\P_{X_o,Y\mid R}(\x_o,y\mid  \rb)\left[Q_{X_m\mid X_o}(\x_m\mid  \x_o)-\P_{X_m\mid X_o,Y, R}(\x_m\mid  \x_o,y, \rb)\right]
	\stepcounter{equation}\tag{\theequation}\label{eq:back}
\end{align*}
Equation \ref{eq:back} suggests it is possible to preserve the joint distribution of $X$ and $Y$, if the imputation model corresponding to the missingness pattern $\rb$, exactly matches the underlying distribution of the missing values conditioned on the observed explanatory variables, response, and the missingness pattern i.e., $\P_{X_m\mid X_o,Y, R}$.  We refer to this latter distribution as the \emph{true missing data conditional distribution} and note that imputing via this conditional distribution is not feasible in the model-X setting as it implicitly relies on the response conditional distribution which is unknown. In turn, imputing without the response or from an inaccurate model can introduce a shifted joint distribution $\P_{\Xh,Y}$ and potentially create an alternative null hypothesis set. We denote this potentially different hypothesis set as follows:
\begin{equation}
	\mathcal{H}_0^{\text{(imp)}}=\{j:Y\indep \Xh_j \mid \Xh_{-j}\}
\end{equation}
%this implies the FDR control model-X provides will be w.r.t to a 
%This is important because an 
%Equation \ref{eq:back} also implies 
%. However, this would require $$
%Although, the ideal imputation requires using both explanatory variables and the response \cite{bibid}, we restrict the imputation to former set inline with the model-X framework which treats conditional distribution unknown. 
To understand how missing data imputation can alter the null hypothesis set, consider an example with two explanatory variables $X_1,X_2$ and a response $Y$ whose joint distribution factors as ${\P_{Y,X_1,X_2}=\P_{Y\mid X_1}\P_{X_2\mid X_1}\P_{X_1}}$. According to this joint distribution, which is visualized in Figure \ref{fig:egmiss}(a), the only null variable is $X_2$ as it satisfies the condition $Y\indep X_2 \mid X_1$. When $X_1$ is observed but $X_2$ is missing (see Figure \ref{fig:egmiss}(c)), imputing the variable $X_2$ using $X_1$ alone does not change its status of being a null variable. However, when $X_1$ is missing but $X_2$ is observed, (see Figure \ref{fig:egmiss}(b)), the imputation relying only on $X_2$, induces a spurious correlation between the originally null variable, $X_2$ and $Y$. Ultimately, when model-X is applied to a dataset resulting from the described process, because the dataset will contain rows where $X_2$ appears as a non-null variable, the $X_2$ would not be treated as a false discovery, and the FDR guarantees of model-X knockoffs will be invalidated. A related example was given in \cite{seijo-pardoBiasesFeatureSelection2019} for illustrating the effects of imputation on univariate statistical tests.

\begin{figure*}[tb]
	\centering
	\subfloat[\centering Both $X_1$ and $X_2$ are observed.]{		
		\begin{tikzpicture}[auto, node distance=1.1cm,every loop/.style={},
	thick,main/.style={circle,draw,minimum size=7mm,inner sep=0pt}]
	
	%			\node[main] (u1)[draw=none] at (0, 0) [rotate=90]{\dots};
	\node[main](3) [] {$X_1$};
	\node[main](4) [below of=3,left of=3] {$Y$};
	\node[main](5) [below of=3,right of=3,above of=4,right of=4] {$X_2$};
	\node[main](6) [below of=5,draw=none] {};	
	% \node[draw=none, text width=2cm, align=center] [above of=u1] {Descendant masking};
	
	\draw [->] (3) to (4);
	\draw [->] (3) to (5);
%	
%	%    	\draw [->](3) to (4);
%	\draw [->,dashed] (u1) to (1);
%	\draw [->,dashed] (u2) to (2);
%	\draw [->,dashed] (4) to (l4);
%	\draw [->,dashed] (5) to (l5);
%	\draw [->,dashed] (6) to (l6);
	%				\node[draw,blue,inner sep=0.1mm,fit=(l4) (3) (l6) (3)] {};
\end{tikzpicture} 
	}
	\hspace*{4em}
		\subfloat[\centering $X_1$ is missing.]{		
		\begin{tikzpicture}[auto, node distance=1.1cm,every loop/.style={},
	thick,main/.style={circle,draw,minimum size=7mm,inner sep=0pt}]
	
	%			\node[main] (u1)[draw=none] at (0, 0) [rotate=90]{\dots};
	\node[main](3) [draw=none] {};
	\node[main](4) [below of=3,left of=3] {$Y$};
	\node[main](5) [below of=3,right of=3,above of=4,right of=4] {$X_2$};
	\node[main](6) [below of=5,right of=4] {$\hat{X_1}$};	
	\node[main](7) [below of=5,right of=6,above of=6,draw=none] {};	
	% \node[draw=none, text width=2cm, align=center] [above of=u1] {Descendant masking};
	
%	\draw [->] (3) to (4);
	\draw [<->] (4) to (5);
		\draw [->] (5) to (6);
	%	
	%	%    	\draw [->](3) to (4);
	%	\draw [->,dashed] (u1) to (1);
	%	\draw [->,dashed] (u2) to (2);
	%	\draw [->,dashed] (4) to (l4);
	%	\draw [->,dashed] (5) to (l5);
	%	\draw [->,dashed] (6) to (l6);
	%				\node[draw,blue,inner sep=0.1mm,fit=(l4) (3) (l6) (3)] {};
\end{tikzpicture} 
	}
	\hspace*{4em}
		\subfloat[\centering $X_2$ is missing.]{		
		\begin{tikzpicture}[auto, node distance=1.1cm,every loop/.style={},
	thick,main/.style={circle,draw,minimum size=7mm,inner sep=0pt}]
	
	%			\node[main] (u1)[draw=none] at (0, 0) [rotate=90]{\dots};
		\node[main](3) [] {$X_1$};
	\node[main](4) [below of=3,left of=3] {$Y$};
	\node[main](5) [below of=3,right of=3,above of=4,right of=4,draw=none] {};
	\node[main](6) [below of=5,right of=4,draw=none] {};	
	\node[main](7) [below of=5,right of=6,above of=6] {$\hat{X_2}$};	
	% \node[draw=none, text width=2cm, align=center] [above of=u1] {Descendant masking};
	
	\draw [->] (3) to (4);
%	\draw [<->] (4) to (5);
%		\draw [->] (5) to (6);
				\draw [->] (3) to (7);
	%	
	%	%    	\draw [->](3) to (4);
	%	\draw [->,dashed] (u1) to (1);
	%	\draw [->,dashed] (u2) to (2);
	%	\draw [->,dashed] (4) to (l4);
	%	\draw [->,dashed] (5) to (l5);
	%	\draw [->,dashed] (6) to (l6);
	%				\node[draw,blue,inner sep=0.1mm,fit=(l4) (3) (l6) (3)] {};
\end{tikzpicture} 
	}
	\hspace*{4em}
	\caption{Example effects of missing data imputation on the null variables. (A) Both $X_1$ and $X_2$ are observed, $X_2$ is a null variable. (B) $X_1$ is missing it and because imputation does not adjust for $Y$, $X_2$ appears as a non-null variable. (C) $X_2$ is missing and because it was originally a null variable, imputed $\hat{X_2}$ remains as null.}
	\label{fig:egmiss}
\end{figure*}

% Despite the obvious effect of imputation, the previous work in model-X literature doesn't  literature model-X

\section{Extending the Model-X Framework to Missing Data}\label{a:sec:main}
% \input{tex/proof_mar}

% In this work, we will introduce different ways for dealing with missing data under the knockoffs framework. Throughout this work, we denote $X=(X_1,\dots,X_p)^T$ as the explanatory R.Vs and $R=(R_1,\dots,R_p)^T$ denote the missing value indicator R.Vs. $R_i$ takes values in $\{0,1\}$ and $R_i=1$ denotes $X_i$ is missing. Let $m=\{i:R_i=1\}$ and $o=\{1,\dots,p\}\setminus m$ denote the sets of missing and observed values. Given a set $S\subseteq\{1,\dots,p\}$, $X_S=\{X_i:i\in S\}$ denotes the R.Vs indexed by set S.

In Section \ref{sec:b_background} we have shown that there are two main pathways missing data violates the assumptions of the model-X framework: shifting the null hypothesis set, and shifting the explanatory variable distribution. The former directly invalidates the FDR control while the latter indirectly does so by violating the pairwise exchangeability. In this section we introduce conditions for preserving the assumptions of model-X when handling the missing entries using CCA and imputation strategies. 

%Just as valid p-values requires the assumptions of the statistical tests to be justified, the assumptions of the model-X knockoffs needs to be justified to control the FDR. As illustrated in Section xx, handling missing data either using CCA or imputation can add further complications. 
%iii) breaking the independent rows.
%ii) Praticallity, using existing knockoff samplers , distribution assumptions might fail 

%Instead, we will show that it's possible to do imputation without making additional assumptions. The idea is because the distribution of $X$ needs to be known for model-X knockoffs, there is no reason for not using $\P_X$ for imputation as well.

% Model-x framework is based on the idea that we have an accurate model for  also determines the optimal imputation algorithm.

%\begin{prop}\label{}
%	If assume 
%	then the resulting joint distribution of the variables and response will be preserved i.e. $P_{X,Y}=P_{X,Y\mid R=\mathbf{1}}$.
%\end{prop}

\subsection{Complete-Case Analysis}

It is well established in the literature that CCA can be biased if  the completely observed examples do not constitute a random sample of the original dataset \cite{littleStatisticalAnalysisMissing2002}. In turn, CCA is unbiased if the missingness pattern is independent of all variables and the response, i.e.
\begin{equation}
	X,Y \indep R.
\end{equation}
%\end{definition}
%This result is already known in the literature in general contexts as discarding under MCAR \cite{bibid} and we included for completeness. This states if the MCAR condition holds the null hypthosis set 
This condition\footnote{Notice, our formulation is defined with respect to the true missingness mechanism.} is referred to as ``everywhere MCAR'' in \cite{seamanWhatMeantMissing2013} and for brevity we will call it MCAR. Under MCAR, because a row's missingness is independent of its values, the completely observed rows are representative of the true distribution of the dataset. This is formalized in the next proposition for completeness. 
%explanatory variables, 
\begin{proposition}\label{a:prop:mcar}
	Assume the MCAR condition holds, and completely observing all explanatory variables has non-zero probability, $\P_{R}(\mathbf{1})>0$, then the joint distribution of the modeled variables, $X$, and $Y$ conditioned on the event that all explanatory variables are completely observed, equals to their unconditional joint distribution, i.e.,
%	equals the unconditional joint distribution of the modeled variables, i.e.,	
	\[\P_{X,Y\mid R}(\x,y\mid \mathbf{1})=\P_{X,Y}(\x,y).\]
\end{proposition}
%$\P_{X,Y}=\P_{X,Y\mid R=\mathbf{1}}$.
\begin{proof}
%	\ref{xx}
	The MCAR condition, $\ X,Y \indep R\ $, implies that the missingness pattern is independent of the modeled variables for all missingness patterns $\rb$ with non-zero probability $\P_{R}(\rb)>0$, i.e.,
	\[\P_{X,Y\mid R}(\x,y\mid \rb)=\P_{X,Y}(\x,y)\]
	for all $(\x,y)\in \mathcal{X}\times \mathcal{Y}$. This indicates, assuming the event $\{R=\mathbf{1}\}$ has non-zero probability, as claimed, the distribution of the completely observed data is equal to the original distribution, i.e., \[\P_{X,Y\mid R}(\x,y\mid \mathbf{1})=\P_{X,Y}(\x,y).\]	
%	 $X,Y \indep R=\mathbf{\rb}$. 
\end{proof}
The practical implication of Proposition \ref{a:prop:mcar} is that the model-X framework remains unbiased under MCAR; that is, existing knockoff samplers satisfies \pety and the null hypothesis set remains unchanged. However, MCAR is a restrictive condition because in practice the missingness mechanism depends on at least some of the modeled variables. Next, we introduce the conditions for using missing data imputation to circumvent the MCAR assumption.

%A less restrictive version MAR assumption xx holds. Next, we discuss how the MAR assumption can be used in the missing data imputation setting.

%conditioned on the all remaining variables (i.e. observed explanatory variables, response and the missingness pattern)
\subsection{Missing Data Imputation }
As shown in section \ref{sec:b_background}, missing data imputation without using the \emph{true missing data conditional distribution}, in general, shifts the joint distribution of the imputed explanatory variables $\Xh$ and the response $Y$ away from the true distribution. Deriving this conditional distribution relies on three separate distributions: the generative model of the explanatory variables, $\P_X$, the response conditional distribution, $\P_{Y\mid X}$, and the missingness mechanism, $\P_{R\mid X,Y}$. In general, the missingness mechanism and the response conditional distribution are unknown, but the generative model is known in the model-X framework. As a remedy, we will show that, for a specific class of missingness mechanisms, using this generative model alone to impute the missing entries can be sufficient to preserve the joint distribution between the explanatory variables and the response.

%the missingness mechanism is ignorable and 
Of the two unknown distributions, it is easier to mitigate the fact that the missingness mechanism is unknown because the assumptions for when missingness mechanisms become ignorable has been established in the literature \cite{littleStatisticalAnalysisMissing2002}. Specifically, when the missingness mechanism satisfies the Missing At Random (MAR) condition, the \emph{true missing data conditional distribution}, no longer depends on the missingness mechanism. 
%The 
%MAR condition in turn holds if, for each probable missingness pattern $\rb$, the missing variables are independent of the missingness pattern being $\rb$ given the observed variables and the response, i.e.,
%\[X_m \indep R=\rb \mid X_o, Y\]
The MAR condition holds, if for all missingness patterns $\rb$
with non-zero probability, $\P_\rvr(\rb)>0$, the missing variables, $X_m$, are conditionally independent of the event $\{R=\rb\}$, given the observed variables $X_o$, i.e.,
\begin{equation}\label{3:eq:mar}
	\P_{X_m \mid R, X_o,Y}(\x_m \mid \rb, \x_o, y)=\P_{X_m \mid X_o, Y}(\x_m \mid \x_o, y),
\end{equation}
holds for all $(\x,y)\in \mathcal{X}\times \mathcal{Y}$ such that $\P_{X_o,Y,R}(\x_o,y,\rb)>0$. If the event that all explanatory variables are missing simultaneously has non-zero probability, i.e., $P_R(\mathbf{0})>0$, then the above \eqref{3:eq:mar} requires ${\P_{X\mid Y, R}(\x\mid y,\mathbf{0})=\P_{X\mid Y}(\x,y)}$ to hold. 

Note that our definition of MAR is with respect to the true missingness mechanism and corresponds to the``everywhere MAR'' definition given in \cite{seamanWhatMeantMissing2013}. Regardless, ignorability of the missingness mechanism, which our definition of MAR implies\footnote{In general an additional conditioned referred to as ``parameter distintness'' is also required \cite{littleStatisticalAnalysisMissing2002}}, is a common assumption which according to the claim given in \cite{molenberghsHandbookMissingData2014} is made by 99\% of practitioners in the context of missing data imputation. 

Mitigating the unknown response conditional distribution requires further restrictions on the missingness mechanism. As one possible solution, we introduce the Missingness with Ignorable Response (MIR) condition. MIR holds if for each probable missingness pattern $\rb$ with $\P_\rvr(\rb)>0$, the missing variables are conditionally independent of the response given the observed variables, i.e.
\begin{equation}X_m \indep Y \mid X_o.
\end{equation}
%We refer to this condition as 
%my themselves does not enable xx but when
These two conditions, MAR and MIR, when combined imply that the missing variables are independent of both the missingness mechanism and the response given the observed entries. We formalize this statement in the next Proposition.
\begin{proposition}\label{prop:a}
	Assume the MAR and MIR conditions hold, then, for each missingness pattern, $\rb$, with non-zero probability, $\P_{R}(\rb)>0$, the variables with missing values $X_m$ are conditionally independent of the response $Y$ and the event $\{R=\rb\}$ given the observed variables, $X_o$ i.e.,
%	\[X_m \indep Y, R=\rb \mid X_o\]

	\[\P_{X_m \mid X_o,Y,R}(\x_m \mid \x_o, y,\rb)=\P_{X_m \mid X_o}(\x_m \mid \x_o),\]
holds for all $(\x,y)\in \mathcal{X}\times \mathcal{Y}$ such that $\P_{X_o,Y,R}(\x_o,y,\rb)>0$.

	%	Let $\Xh=(\Xh_1,\dots,\Xh_p)^T$ denote the imputed vector where $\Xh_o=X_o$. If $\hat{X}_m\sim P_{X_m\mid X_o,R}(\cdot;X_o,R)$ and then $P_X=P_{\Xh}$.
\end{proposition}
%	MIR and MAR conditions can be combined xx.
\begin{proof}[Proof of Proposition \ref{prop:a}]
	MAR condition (\eqref{3:eq:mar}) states for all probable missingness patterns $\rb$, the missing variables are conditionally independent of the event $\{R=\rb\}$, given the observed variables, i.e.,
\begin{align*}
			\P_{X_m \mid X_o,Y,R}(\x_m \mid \x_o, y,\rb)&=\P_{X_m \mid X_o, Y}(\x_m \mid \x_o, y)	\stepcounter{equation}\tag{\theequation}\label{3:eq:proof_a}\\
	\intertext{for all $(\x,y)\in \mathcal{X}\times \mathcal{Y}$ such that $\P_{X_o,Y,R}(\x_o,y,\rb)>0$. In turn MIR condition states for each probable pattern $\rb$, the corresponding missing variables are conditionally indepedent of the response given the observed variables, i.e.,}
	\P_{X_m \mid X_o, Y}(\x_m \mid \x_o, y)&=\P_{X_m \mid X_o}(\x_m \mid \x_o) 
%	\intertext{}
	\stepcounter{equation}\tag{\theequation}\label{3:eq:proof_b}\\
	\intertext{for all $(\x,y)\in \mathcal{X}\times \mathcal{Y}$ such that $\P_{X_o,Y}(\x_o,y)>0$. This implies
		\eqref{3:eq:proof_b} holds for all $(\x,y)\in \mathcal{X}\times \mathcal{Y}$ such that $\P_{X_o,Y,R}(\x_o,y,\rb)>0$. In turn, applying equations \ref{3:eq:proof_a} and \ref{3:eq:proof_b} sequentially implies the main statement of Proposition \ref{prop:a} indeed holds}
		\P_{X_m \mid X_o,Y,R}(\x_m \mid \x_o, y,\rb)&=\P_{X_m \mid X_o, Y}(\x_m \mid \x_o, y)=\P_{X_m \mid X_o}(\x_m \mid \x_o)
\end{align*}
%Because any tuple $(\x,y)$ that satisfies $\P_{X_o,Y,R}(\x_o,y,\rb)>0$ also satisfies $\P_{X_o,Y}(\x_o,y)>0$, 
for all $(\x,y)\in \mathcal{X}\times \mathcal{Y}$ such that $\P_{X_o,Y,R}(\x_o,y,\rb)>0$. This proof essentially uses the contraction property of conditional independence statements
\cite{pearlCausalityModelsReasoning2000} at the event level to combine the MAR and MIR conditions.
\end{proof}

Proposition \ref{prop:a} has an important implication. Because the missing variables are independent of the response and the missingness mechanism's realization when conditioned on the observed variables, the \emph{true missing data conditional distribution} no longer relies on the two unknown distributions: the missingness mechanism and the response conditional distribution. This enables using only the known generative model, $\P_X$ while imputing and still preserving the original distribution of the modeled variables, as formalized in the next Proposition.

\begin{proposition}\label{prop:b}
	Assume the MAR and MIR conditions hold. If for each missingness pattern $\rb$ with non-zero probability, $\P_{R}(\rb)>0$, the imputation distribution matches the corresponding conditional distribution of the missing variables given the observed variables, i.e.,
	\[Q_{X_m\mid X_o}=\P_{X_m\mid X_o},\]
	then the resulting joint distribution of the imputed variables and the response equals the original joint distribution of the modeled variables, i.e., $\P_{\Xh,Y}=\P_{X,Y}$.
\end{proposition}
\begin{proof}[Proof of Proposition \ref{prop:b}]
	To show Proposition \ref{prop:b} holds, we refer to \eqref{eq:back} in Section \ref{sec:b_background}, which states the difference between the original joint distribution, $\P_{X,Y}$, and the joint distribution resulting from imputation, $\P_{\Xh,Y}$, is given by
	\begin{align*}
		\P_{\hat{X},Y}(\x,y)&-\P_{X,Y}(\x,y)\\
		&=\sum_{\rb\neq\mathbf{1}}\P_R(\rb)\P_{X_o,Y\mid R}(\x_o,y\mid  \rb)\left[Q_{X_m\mid X_o}(\x_m\mid  \x_o)-\P_{X_m\mid X_o,Y, R}(\x_m\mid  \x_o,y, \rb)	\right]\\
		\intertext{Because the MAR and MIR conditions hold, Proposition \ref{prop:a} applies. Accordingly, for all probable $\rb$, the missing variables are conditionally independent of the response and the event $\{R=\rb\}$ given the observed variables. As the summation is implicility applied over the probable $\rb$, Proposition 2 can be used to obtain:}
		\P_{\hat{X},Y}(\x,y)&-\P_{X,Y}(\x,y)\\	&=\sum_{\rb\neq\mathbf{1}}\P_R(\rb)\P_{X_o,Y\mid R}(\x_o,y\mid  \rb)\left[Q_{X_m\mid X_o}(\x_m\mid  \x_o)-\P_{X_m\mid X_o}(\x_m\mid  \x_o)
		\right]\\
		\intertext{Plugging in the assumption of Proposition \ref{prop:b}, which states the imputation model matches the conditional distribution of the missing variables given the observed explanatory variables, i.e. $Q_{X_m\mid X_o}=P_{X_m\mid X_o}$}
		\P_{\hat{X},Y}(\x,y)&-\P_{X,Y}(\x,y)=0
		\stepcounter{equation}\tag{\theequation}\label{}
	\end{align*}
	We have showed that when both the MAR and MIR conditions hold and the imputation model satisfies $Q_{X_m\mid X_o}=\P_{X_m\mid X_o}$,
	then, as claimed in Proposition \ref{prop:b}, the resulting joint distribution of the variables and response is preserved, i.e., $\P_{X,Y}=\P_{\Xh,Y}$
	%, which implies $P_{X,Y}=P_{\Xh,Y}$ holds if the imputation model exactly matching the underlying distribution of the missing values conditioned on the observed explanatory variables, response and the missingness pattern $\P_{X_m\mid X_o,Y, R=\rb}$.

	%$Q_{X_m\mid X_o}=P_{X_m\mid X_o}=\P_{X_m\mid X_o,Y, R}$ holds.
	
	%\[\sum_{\rb}\P_R(\rb)\P_{X_o,Y\mid R}(\x_o,y\mid  \rb)\left[Q_{X_m\mid X_o}(\x_m\mid  \x_o)-\P_{X_m\mid X_o,Y, R}(\x_m\mid  \x_o,y, \rb)\]
	
\end{proof}

Proposition \ref{prop:b} states imputation using only the generative model is sufficient to prevent data distribution shift when the missingness mechanism satisfies MAR and MIR. However, the question whether these two assumptions are reasonable assumption in practice remains. Differentiating between the MAR and MNAR is known to be challenging \cite[S15.1]{molenberghsHandbookMissingData2014}, potentially due to the non-idenfiability of the missingenss mechanism \cite{molenberghsEveryMissingnessNot2008}. Similar challenges also face the MIR assumption as the response conditional distribution is unknown. 
%is known to be untestable in gnere
%First, consider the missingness mechanism, even the missingness mechanism is unknown, it is when it satifies the ignorability condition 
%the missing variables are independent of the missingness pattern conditioned on the observed variables (and the response), the missiness mechanism becomes ignorable \cite{rubin}. This condition is referred to as MAR and can be defined formally using 
%\begin{definition}
%	MAR condition holds if for all probable missingness pattern $\rb$, the missing variables are independent of the missingness pattern being $\rb$ given the observed variables and the response, i.e.
%	\[X_m \indep R=\rb \mid X_o, Y\]
%\end{definition},
%This definition essentially follows the ``everywhere MAR'' definition given in \cite{seaman}. MAR condition is also referred to as ``ignorable missingness'' because, while computing the \emph{true missing data conditional distribution}, ignoring the missingness mechanism \cite{bibid}. 
%MCAR condition, we considered for the CCA has to satisfy the MAR but not vice versa. 
%In practise of the satisfy the MIR condition???

To conceptualize when MIR holds, we provide a concrete scenario. Assume the response variable $Y$, has a Markov Blanket, \cite{margaritisProvablyCorrectFeature}, i.e. a set $\mathcal{M}\subseteq\{1,\dots,d\}$, such that all variables outside the Markov Blanket are conditionally independent of the response given the variables in the blanket:
\begin{equation}Y\indep X_{\overline{\mathcal{M}}} \mid X_\mathcal{M}.
\end{equation}
In the next proposition we show that when the set of variables in the Markov Blanket of $Y$ do not contain any missing data, the MIR condition holds.
 
% This set is a less restricted definition of the Markov Boundary refers to the smallest Markov Blanket. 
 
% and this set is always observed. Such a set is referred to as Markov Blanket in the literature \cite{bibid} and formally defined as follows: 
% \begin{definition}
% 	Let $\mathcal{M}\subseteq\{1,\dots,p\}$. $X_{\mathcal{M}}$ is the Markov Blanket of $Y$ if xx
% 	\[Y\indep X_{\overline{\mathcal{M}}} \mid X_\mathcal{M}\]
% 	holds.
% \end{definition}

\begin{proposition}\label{a:prop:mb}
	If there exists a set $\mathcal{M}$ which is a Markov Blanket of Y, and is always observed, i.e., for all missingness patterns $\rb$ with non-zero probability, the observed entries contain the set $\mathcal{M}$, i.e. $o \supseteq \mathcal{M}$, then the MIR condition holds.
\end{proposition}
\noindent Before we begin the proof of Proposition \ref{a:prop:mb}, we will restate the weak union property of conditional independence statements given in \cite{pearlCausalityModelsReasoning2000}:
\begin{lemma}[Weak union property given in \cite{pearlCausalityModelsReasoning2000}]\label{a:lemma:2} Given four random vectors $X$, $Y$, $W$, and $Z$, the weak union property states
		\begin{equation*}
			X \indep Y,W \mid Z \implies X\indep Y \mid Z,W
		\end{equation*}
\end{lemma}
\noindent We next proceed to the proof of Proposition \ref{a:prop:mb}.
\begin{proof}[Proof of Proposition \ref{a:prop:mb}]	
	To show the Proposition holds, we will start with the definition of the Markov Blanket $\mathcal{M}$ is a set such that
	\begin{align*}
		&X_{\overline{\mathcal{M}}} \indep  Y \mid X_\mathcal{M}
		\intertext{Because this Markov Blanket of Y is always observed, it is contained within the observed entries $o$ for all probable $\rb$. This fact, combined with the observed, $o$, and missing indices, $m$, forming a partition, results in the variables outside of the blanket splitting as follows:}
		&X_{\overline{\mathcal{M}}} \indep  Y \mid X_\mathcal{M}\iff X_{m},X_{o\setminus \mathcal{M}} \indep Y \mid X_\mathcal{M}\\
		\intertext{Here, the weak union property of conditional independence statements (Lemma \ref{a:lemma:2}) imply the missing variables are independent of the response conditioned on the variables which are in the Markov Blanket of Y and the variables which are observed but outside of the blanket. }
		&X_{m},X_{o\setminus \mathcal{M}} \indep Y \mid X_\mathcal{M} \implies X_m \indep Y \mid X_\mathcal{M}, X_{o\setminus \mathcal{M}}
		\intertext{Because $\mathcal{M}$ is always observed, the two sets, $\mathcal{M}$ and $o\setminus \mathcal{M}$, form a partition of the observed variables $o$.}
		&X_m \indep Y \mid X_\mathcal{M}, X_{o\setminus \mathcal{M}} \iff X_m \indep Y \mid X_{o} \stepcounter{equation}\tag{\theequation}\label{a:eq:proof}
	\end{align*}
		Because \eqref{a:eq:proof} holds for all probable $\rb$, as claimed, when there exists a Markov Blanket of Y that is always observed, the MIR property holds also holds for all probable $\rb$.
\end{proof}
%X_m \indep Y \mid X_{o}
Although Proposition \ref{a:prop:mb}, is helpful to conceptualize when the MIR condition holds, it also in ways can be considered contradictory, as the Proposition implies the variables containing missing data are not informative of the response (outside of the Markov Boundary) which justifies omitting the variables containing missing data from the analysis completely.

%$\{x^{(i)}_{o^{i}}\}_{i=1}^N$ with varying missing values indices $\{m^{i},o^{i}\}_{i=1}^N$. 
Regardless, given that we proved conditional imputation preserves the data distribution, under the MAR and MIR conditions, in Algorithm \ref{alg:main} we illustrate how it can be applied to a set of N i.i.d. observations. The only additional step introduced in Algorithm \ref{alg:main} over the standard model-X framework is the sampling of missing values from the conditional distribution.
%\todo[inline]{define $o^i$}
% Because of the i.i.d. assumption the resulting imputed matrix and knockoffs matrix will be pairwise exchangeable as well. 

% given N observations $\{x^{(i)}_{o^{i}}\}_{i=1}^N$ 
% \begin{algorithm}[H]
%\begin{algorithm}[H]
%	\caption{Imputing the missing values by conditional sampling and subsequently sampling the knockoffs}\label{alg:main}
%	%   \textbf{Input:}  $\{x^{(i)}_{o^{i}}\}_{i=1}^N, \{m^{i},o^{i}\}_{i=1}^N, P_{\Xt\mid X}, P_X$
%	\begin{algorithmic}
%		\STATE {\bfseries Input:} $\{\x^{(i)}_{o^{i}}\}_{i=1}^N, \{m^{i},o^{i}\}_{i=1}^N, \P_{\Xt\mid X}, \P_X$
%		\FOR{\texttt{$i=1,\dots,N$}}
%		\STATE $m\leftarrow m^i$
%		\STATE $o\leftarrow o^i$
%		\STATE $\hat{\x}_{m}\sim \P_{X_{m}\mid X_{o}}(\cdot;\x^{(i)}_{o})$ \hfill\COMMENT{Posterior Imputation}
%		\STATE  $\hat{\x}^{(i)}\leftarrow(\x^{(i)}_{o},\hat{\x}_{m})$
%		\STATE$\xt^{(i)} \sim \P_{\Xt\mid X}(\cdot;\hat{x}^{(i)}) $\hfill\COMMENT{Knockoff Sampling}
%		\ENDFOR
%		\STATE {\bfseries Output:} $\{\hat{x}^{(i)},\xt^{(i)}\}_{i=1}^N$
%	\end{algorithmic}
%\end{algorithm}

\begin{algorithm}[H]
	\caption{Imputing the missing values by conditional sampling and subsequently sampling the knockoffs}\label{alg:main}
	%   \textbf{Input:}  $\{x^{(i)}_{o^{i}}\}_{i=1}^N, \{m^{i},o^{i}\}_{i=1}^N, P_{\Xt\mid X}, P_X$
	\begin{algorithmic}
		\STATE {\bfseries Input:} $\Xbr, \ \P_{\Xt\mid X}, \ \P_X$
		\STATE $\Xbh \leftarrow \Xbr$\hfill\COMMENT{Initialize the $N\times d$ imputation matrix}
		\STATE $\Xtb \leftarrow \mathbf{0}$ \hfill\COMMENT{Initialize the $N\times d$ knockoff matrix}
		\FOR{\texttt{$i=1,\dots,N$}}
		\STATE $m\leftarrow \{j:\Xbr_{i,j}=\text{NA}\}$
		\STATE $o\leftarrow \{j:\Xbr_{i,j}\neq\text{NA}\}$
%		\STATE $\xh^{(i)} \leftarrow \mathbf{0}$ \hfill\COMMENT{Initialize}
%		\STATE  $\xh_o\leftarrow \Xbh_{i,o}$
		\STATE $\Xbh_{i,m}\sim \P_{X_{m}\mid X_{o}}(\cdot;\Xbh_{i,o})$ \hfill\COMMENT{Sampling from the conditional distribution}
%		\STATE $\Xbh_{i,m}\leftarrow \xh_{m}$
%		\STATE  $\hat{\x}^{(i)}\leftarrow(\x^{(i)}_{o},\hat{\x}_{m})$
		\STATE$\Xtb_i \sim \P_{\Xt\mid X}(\cdot;\Xbh_{i}) $\hfill\COMMENT{Sampling the knockoffs}
		\ENDFOR
		\STATE {\bfseries Output:} $\Xbh,\Xtb$
	\end{algorithmic}
\end{algorithm}

\section{Simulation Experiments}\label{a:sec:results}

% To investigate the effect of missing values in the knockoffs framework, we conducted various simulations. 
In this section we provide computational simulations to assess our theoretical findings. Because of the difficulty of obtaining ground truth variables in real world datasets, simulation studies are commonly used in the model-X knockoffs literature \cite{candesPanningGoldModelX2017,barberControllingFalseDiscovery2015a,nguyenAggregationMultipleKnockoffs2020}. 

\subsection{Experimental Setup}
In our experiments, we have investigated the effect of missing values in the knockoffs framework by simulating a feature selection problem in different settings. 

%The simulations are divided into two categories based on the probability distribution of the explanatory variables, namely multivariate normal (MVN) distributed.

% varied the number of missing values, the correlation among the variables, and the missing data pattern. 

% Moreover in this setting, we also compared our imputation methods with methods that do not come with FDR guarantees. Finally, we observed the joint effect of changing the number of samples and the number of missing values when the data is distributed according to an HMM. 

\paragraph{Simulation Setup, Response Distribution:} Throughout our experiments, we have followed the simulations of \cite{sesiaGeneHuntingHidden2019} and used a logistic regression model as the response conditional distribution. Additionally, we simulated missing data in the explanatory variables. Let $X$ be a d-dimensional random vector with a specified $\P_X$. The response variable is given by:
%\todo[inline]{maybe use the slides notation}
%, \quad \text{\quad }\beta^{*}_i=
%\begin{cases}
%	a,&\text{if } i\in S^*\\0,              &\text{otherwise}
%\end{cases}\]
\begin{equation}Y \mid X \sim \text{Bernoulli}(p=\sigma(\sum_{j\in S^*} \alpha X_j))
\end{equation}
% \[Y\sim \text{Bernoulli}(p=\sigma(X^T\beta^{*}))\]
% where 
% \[\beta^{*}_i=
% \begin{cases}
%     a,&\text{if } i\in S^*\\0,              &\text{otherwise}
% \end{cases}
% \]
where $\sigma$ denotes the logistic function, $\alpha$ denotes the amplitude of the coefficient and $S^*$ denotes the indices of the subset of covariates with non-zero coefficient. The number of variables with non-zero coefficients, i.e. $|S^*|$, is an experiment parameter and elements of $S^*$ are randomly selected from $\{1,\dots,d\}$ with equal probability and kept the same for different trials.

\paragraph{Missingness Mechanism:} After generating N i.i.d samples from $\P_X$ and stacking them as rows of the matrix $\Xb$, we simulate a MCAR distribution with $R_j \sim \text{Bernoulli}(p=p_j)$ where $p_j=1-p_0$ if $j\in \mathcal{R}$ and $p_j=1$ other wise where $\mathcal{R}\subseteq\{1,\dots,d\}$ denotes the variables, for which missing values can occur and $p_0$ denotes the rate of missingness. The resulting partially observed design matrix $\Xbr$ and the response vector $\Yb$ are then used as the input to the knockoffs procedures. 

%The N i.i.d samples of observed variables and the response variables i.e. $\{X_{o^i}^{(i)},Y^{(i)}\}_{i=1}^N$ are then used as the input to the knockoffs procedure. 

% is a simulation parameter, $\epsilon_i \sim \text{Bernoulli}(p=p_0)$, and $\{e_i\}_{i=1}^p$ are i.i.d.

% N i.i.d samples $\{X'^{(i)},Y^{(i)}\}_{i=1}^N$ are drawn after specifying $P_X$ and used as the input to the knockoffs procedure. 
% As the score estimator the absolute value of the coefficients of a logistic regression model with $l_1$ regularization is used. For the multivariate normal data each time $\lambda$ is selected using 5-fold-cross-validation with area under the curve metric and search space $\{1e-10,1e-2,1e-1,1,1e1\}$ is used. For HMM $\lambda$ is fixed to 1. 

\paragraph{Model-x Knockoffs Setup:} In the model-X knockoffs framework one must specify the feature-scorer and the final weighted score of each variable. As the feature-scorer we have used the coefficients of an $\ell1-$regularized logistic regression model \cite{hastieStatisticalLearningSparsity2019} of the concatenated matrix $[\Xbh,\Xtb]$ which optimizes the following objective function:
\begin{equation}\hat{\beta}(\lambda), \hat{\beta}_0(\lambda) = \argmin_{\beta,\beta_0} \sum_{i=1}^N \log(1+e^{-y_i(\beta^T[x^{(i)},\xt^{(i)}]+\beta_0)})+\lambda \norm{\beta}_{1}
\end{equation}

In each trial hyper-parameter $\lambda$ is selected using 5-fold-cross-validation using the area under the curve metric and search space $\{1e-10,1e-2,1e-1,1,1e1\}$. Let $\lambda^*$ denote the resulting optimal hyper-parameter and $\hat{\beta}(\lambda^*)\in \real^{2d}$ denote the estimated coefficients, as common in the literature (see for e.g. \cite{sesiaMultiresolutionLocalizationCausal2020}), the feature scores are determined by the absolute linear weights:
\begin{equation}
	\begin{split}
		T_i&=|\hat{\beta}(\lambda^*)_i|, \quad \text{and}\quad \Tt_i=|\hat{\beta}(\lambda^*)_{i+d}|.
%		W_i &= T_i-\Tt_i, \quad i=1,\dots, p
	\end{split}
\end{equation}
% \clearpage

% \clearpage
\subsection{Experimental Results}
%MVN Simulations: Correlation and Missing Values
% $\Sigma_{ij}=\rho^{|i-j|}$
In these simulations, our goal was to observe how the amount of the missing data and the correlation among the explanatory variables affects the performance of the developed missing value knockoffs method. For that reason, following \cite{barberControllingFalseDiscovery2015a,nguyenAggregationMultipleKnockoffs2020} we have used a zero-mean MVN with correlation matrix $\Sigma$ with entries
$\Sigma_{ij}=\rho^{|i-j|}$. This structure is chosen because it controls the correlations with a single parameter $\rho\in(0,1]$ which we refer to as the \emph{correlation strength}. Another question we had was whether the violated MIR condition effected the performance. Accordingly, we considered two different experimental setups: one in which the missing values are restricted to occur at the true variables ($\mathcal{R}=S^*$) and the MIR condition is violated; and another in which the missing values are restricted to null variables ($\mathcal{R}=\{1,\dots,d\}\setminus S^*$) and the MIR condition holds. In each experiment setting, we have searched the grid of different missingness rates ($p_0=\{0,0.1,\dots,0.4\}$), and correlation strength parameters ($\rho=\{0,0.1,\dots,0.8\}$) to investigate the bivariate relationship. 

%While in the third experimental setting we didn't restrict the location ($M=\{1,\dots,p\}$). 

% In this simulation category we have used a MVN distribution for the explanatory variables i.e. $X\sim N(0,\Sigma)$. Following \cite{barberControllingFalseDiscovery2015a,nguyenAggregationMultipleKnockoffs2020} to probe the effect of correlation on the performance we have parametrized the covariance matrix s.t. the correlation between two variables gets lower as their positional distance increase i.e. $\Sigma_{ij}=\rho^{|i-j|}$ where $\rho\in(0,1]$ controls the correlation amount. 

% In the MVN simulations, we investigated how the correlation of the explanatory variables and the missing data amount affected feature selection. We also explored what happens to the performance when the missing data occurs only in the true variables or only in the nulls variables. Another comparison we made was between the readily available imputation methods that do not guarantee FDR control under the knockoffs framework and the methods we introduced.

% we have used the below multivariate normal (MVN) distribution where .

% \[X\sim N(\mu,\Sigma)\]

% \[\ \text{where} \quad \Sigma= \begin{bmatrix}
% \rho^0 & \rho^1 & \dots & \rho^{p-1}\\
% \rho^1 & \rho^0 & \dots & \rho^{p-2}\\
% \vdots & \vdots & \dots & \vdots\\
% \rho^{p-1} & \rho^{p-2} & \dots & \rho^{0}\\
% \end{bmatrix}, \quad \mu=0 \]

%exp. parameters
Other experimental parameters we selected are as follows: we have used $d=700$ different explanatory variables and set $N=1050$ which is slighly higher than $d$. Six percent of the variables are selected as true, i.e., $\mid S^*\mid =42$ and the effect size is set to $\alpha=10/ \sqrt{N}=0.38$. Each setting in the grid is repeated 31 times to obtain empirical estimates of the False Discovery Proportion (FDP), defined as $\frac{\mid \hat{S} \setminus  S^*\mid}{\mid \hat{S}\mid}$, and power, defined as $\frac{\mid \hat{S}\cap S^*\mid}{\mid S^*\mid}$. For the knockoff samplers required in Algorithm \ref{alg:main} we have used the Model-X MVN knockoffs introduced in \cite{candesPanningGoldModelX2017} which requires solving a convex optimization problem. To sample from the posterior distribution $\P_{X_m\mid X_o}$(as required in Algorithm \ref{alg:main}), we have used the exact inference formulas for the MVN distribution.

%  We have explored three different missing value configurations. The locations where missing values can occur are as follows:

% We have used the experiment parameters $p=700$, $a=0.38$, $\mid S\mid =42$.  but kept $N=1050$. Experiments are repeated 31 times. 

% Posterior probability distributions of MVNs used in Algorithm 1 have a closed-form solution (but it requires matrix inversion).

%different variations

% \input{tex/results_mvn_p700}

% \paragraph{Results}

% and average FDR is generally controlled at the target $q=0.1$ (Fig \ref{fig:mvn3})

\begin{figure}[H]
	% [!htb]
	\centering
		\includegraphics[width=1\linewidth]{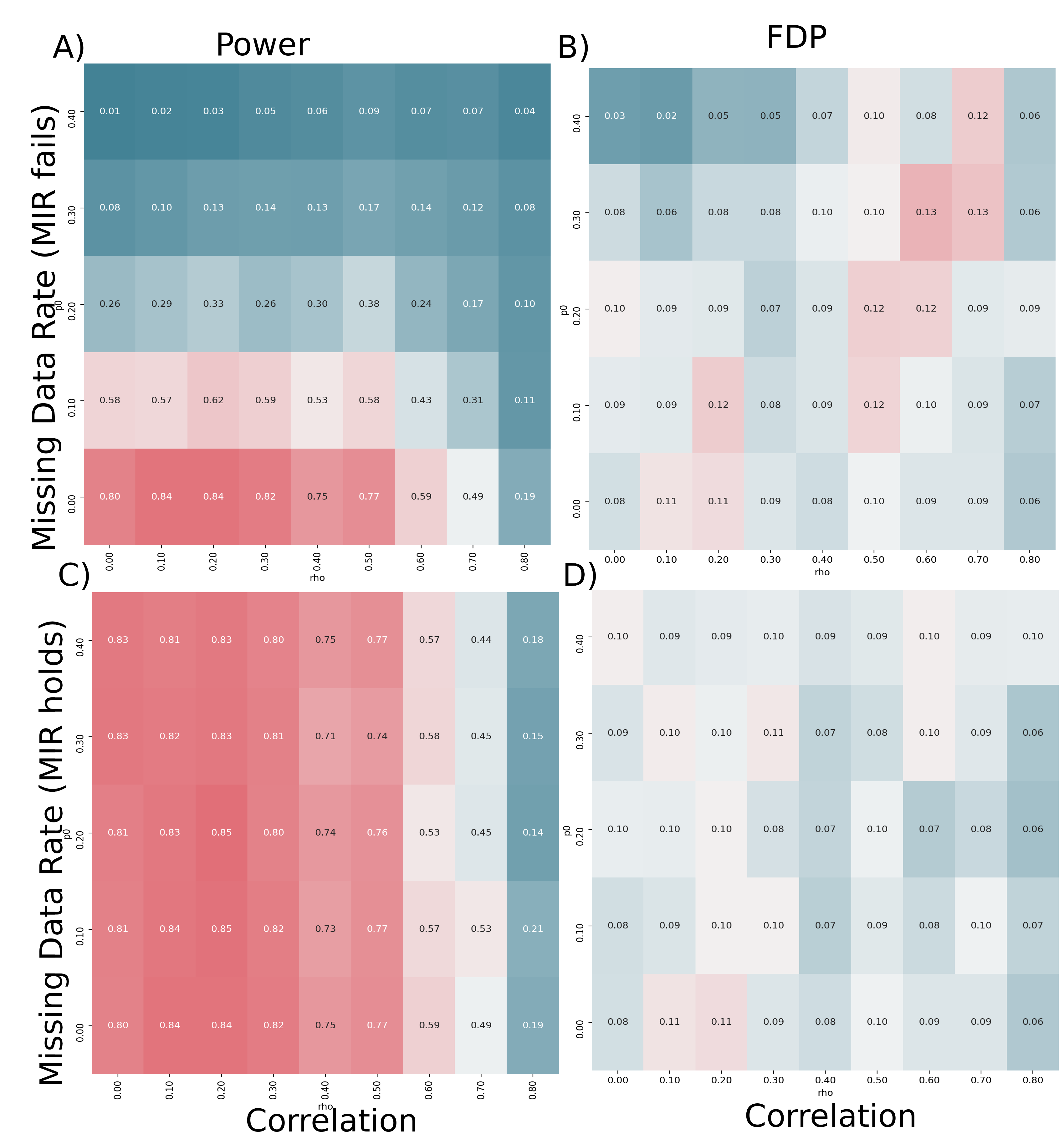}
	\caption{Annotated heatmaps of estimated power (A) \& (C) and FDP (B) \& (D) resulting from model-X with imputation Algorithm \ref{alg:main}. The missing value positions are restricted to true features on the \emph{top row}, to null variables on the \emph{bottom}. X axes vary the correlation among the explanatory variables $\rho$ and y axes denote the missingness rate: $p_0$. For FDP, values above target $q=0.1$ are denoted increasingly red.}
	\label{a:fig:mvn_12}
\end{figure}

After conducting the experiments, we have observed that the FDP is generally controlled at the target $q=0.1$ (see Figure \ref{a:fig:mvn_12}B \& D). Occasionally the FDP peaked above the target level, and this was more likely to occur in the setup when missing values occurred at the true values compared to when missing values were limited to null variables. This can be explained by the violated MIR condition when the true variables contain missing data and accordingly the missing variables still being informative about the response. 

%especially when variables had moderate correlation ($\rho>0.5$)

% This is expected as when the missing values occurring at the true variables invalidates the MIR condition and FDR control of conditional imputation is not guaranteed.

%when the missing values occur at the true variables, we observed the average FDR (out of 31 trials) can reach up to 0.13 when 

% regardless of the missing value pattern, number of missing values, and the correlation amount (Fig \ref{fig:mvn3}). 

We also noticed that as the correlation strength among the explanatory variables $\rho$ increased, the average power consistently decreased in the two missing value configurations (Fig \ref{a:fig:mvn_12} A\&C). On the contrary, the affect of missing data rate on power ($p_0$) was different for the experiments. Specifically, when the missing values were restricted to the true variables and the MIR condition was violated, increasing the missing data rate ($p_0$) decreased the power (Fig \ref{a:fig:mvn_12} A). However, when the missing values were restricted to the nulls and the MIR condition held, the power was mostly unaffected from the missingness rate $p_0$ in (Fig \ref{a:fig:mvn_12}C).

\section{Discussion and Relevant Work}\label{a:sec:rel}
%\section{Relevant Work}
In this work, we studied the effects of missing data on the recently introduced model-X framework, but our study had limitations, and several open problems still remain. One possible future direction is replacing the MIR condition we introduced to preserve the data distribution during imputation with less restive variations. This could be possible by identifying sufficient conditions not for preserving the data distribution but specifically the null hypothesis set. Another future direction could be to explore the assumption that the missingness mechanism is known. This assumption can enable jointly modeling the explanatory variables and missingness variables within the model-X framework and potentially bypassing the imputation step. Finally, our experiments focused only on the MCAR missingness mechanism. It is important to characterize the effects of more general missingness mechanisms on FDR and statistical power.

In our review of the literature, we have observed only one work that discusses the nuances of missing data in the model-X framework \cite{sesiaMultiresolutionLocalizationCausal2020}. This paper, in the context of genome wide association studies, argues imputing \emph{completely} unmeasured latent variables using an existing model can be informative for univariate analysis but not for the multivariate analysis model-X provides. This argument does not directly translate to the missing data setting because, unlike a latent variable, a partially observed variable is still informative in the examples (i.e., rows) it is observed. In practical applications of model-X, we observed missing data is instead either directly imputed or variables containing missing data are completely removed from analysis. For example, in \cite{candesPanningGoldModelX2017, masudUtilizingMachineLearning2021} missing values are first imputed and the Multivariate Normal (MVN) distribution is fit to the resulting imputed matrix. This is problematic, because the imputed matrix may no longer MVN and second, $\P_X$ is estimated from the same data used for feature selection; this introduces dependence between the rows of the knockoff matrix, $\Xtb$. In two other papers, we have observed the variables with the missing entries are completely removed \cite{jiangKnockoffBoostedTree2021,fuHighdimensionalVariableSelection2021}. Although this latter approach does not invalidate the guarantees of the knockoffs framework, this can result in omitting potentially important variables.

\section{Conclusion}

 In this work, we have formalized the effects of missing data on the principled feature selection method, model-X knockoffs. We have shown that imputing from the conditional distribution of the generative model for certain class of missingness mechanism preserves \emph{true missing data conditional distribution}. Next, in our experiments, we characterized the dependence of statistical power and FDR on the correlation strength among the variables and the missing data rate and the necessity of the MIR assumption to impute with the generative model. Overall, our theoretical and experimental findings indicate that missingness impacts the model-X procedure, but theoretical guarantees can be preserved under certain assumptions. 

\bibliography{overall, chap1/sample}
\end{document}